\pdfoutput=1

\documentclass[11pt]{article}

\usepackage{EMNLP2022}

\usepackage{times}
\usepackage{latexsym}
\usepackage{amsmath}
\usepackage{graphicx}
\usepackage{amsfonts}
\usepackage{float}
\usepackage{pgfplots}
\usepackage{booktabs}
\usepackage{multicol}
\usepackage{multirow}
\usepackage{arydshln} 
\usepackage{pgfplots}
\usepackage{subfigure}
\usepackage[T1]{fontenc}
\usepackage[utf8]{inputenc}
\usepackage{microtype}
\usepackage{inconsolata}
\usepackage{enumitem}
\newcommand\blfootnote[1]{%
  \begingroup
  \renewcommand\thefootnote{}\footnote{#1}%
  \addtocounter{footnote}{-1}%
  \endgroup
}

\title{Instance Regularization for Discriminative Language Model Pre-training}

\author{Zhuosheng Zhang\textsuperscript{1,2*}, Hai Zhao\textsuperscript{1,2}, Ming Zhou\textsuperscript{3} \\
  $^1$Department of Computer Science and Engineering,  Shanghai Jiao Tong University\\ 
  $^2$Key Laboratory of Shanghai Education Commission for Intelligent Interaction \\
and Cognitive Engineering, Shanghai Jiao Tong University \\
  $^3$Langboat Technology \\
{\tt{zhangzs@sjtu.edu.cn, zhaohai@cs.sjtu.edu.cn, zhouming@chuangxin.com}} \\}

\begin{document}
\maketitle

\blfootnote{* Work done during internship at Lanboat. This work was partially supported by Key Projects of National Natural Science Foundation of China (U1836222 and 61733011).}

\begin{abstract}
Discriminative pre-trained language models (PrLMs) can be generalized as denoising auto-encoders that work with two procedures, ennoising and denoising. First, an ennoising process corrupts texts with arbitrary noising functions to construct training instances. Then, a denoising language model is trained to restore the corrupted tokens. Existing studies have made progress by optimizing independent strategies of either ennoising or denosing. They treat training instances equally throughout the training process, with little attention on the individual contribution of those instances. To model explicit signals of instance contribution, this work proposes to estimate the complexity of restoring the original sentences from corrupted ones in language model pre-training. The estimations involve the corruption degree in the ennoising data construction process and the prediction confidence in the denoising counterpart. Experimental results on natural language understanding and reading comprehension benchmarks show that our approach improves pre-training efficiency, effectiveness, and robustness. Code is publicly available at \url{https://github.com/cooelf/InstanceReg}

\end{abstract}

\section{Introduction}
Leveraging self-supervised objectives to pre-train language models (PrLMs) on massive unlabeled data has shown success in natural language processing (NLP) \citep{peters2018deep,radford2018improving,devlin-etal-2019-bert,dong2019unified,lan2019albert,clark2019electra,luo2021leveraging,zhu2022leveraging}. A wide landscape of pre-training objectives has been produced, such as autoregressive \citep{radford2018improving,yang2019xlnet} and autoencoding \citep{devlin-etal-2019-bert,joshi-etal-2020-spanbert} language modeling objectives, which serve as the principled mechanisms to teach language models general-purpose knowledge through the pre-training, and then those pre-trained PrLMs can be fine-tuned for downstream tasks. Based on these unsupervised functions, three classes of PrLMs have been proposed: autoregressive language models (e.g. GPT \citep{radford2018improving}), autoencoding models (e.g. BERT \citep{devlin-etal-2019-bert}), and encoder-decoder models (e.g. BART \citep{lewis2020bart} and T5 \citep{raffel2020exploring}). In this work, we focus on the research line of autoencoding models, also known as discriminative PrLMs that have achieved impressive performance on natural language understanding (NLU).

\begin{figure}
	\centering
	\includegraphics[width=0.48\textwidth]{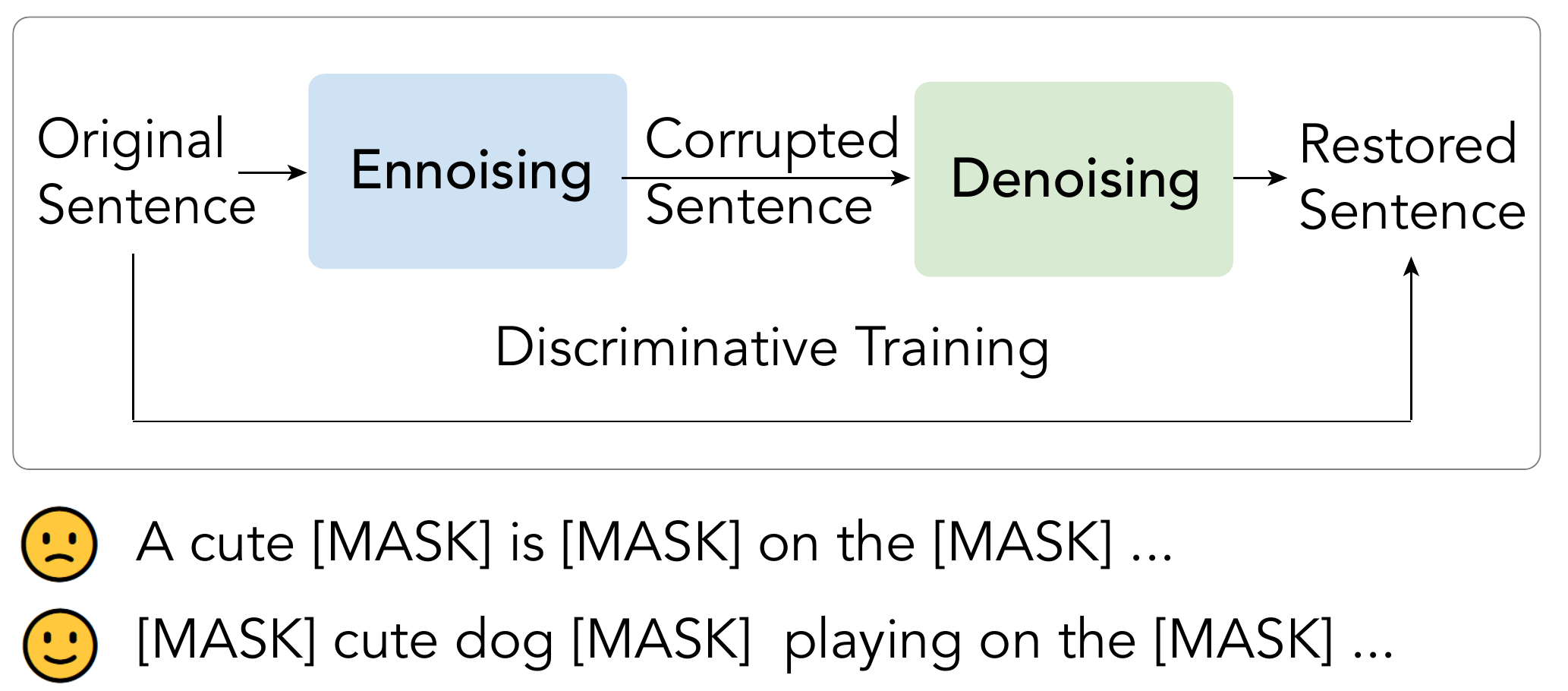}
	\caption{Overview of AutoDecoders. As the two examples show, the random sampling operation during ennoising would result in training instances of different degree of difficulty, e.g., the variety of valid alternatives.}
	\label{fig:autodecoder}
\end{figure}

Although the discriminative PrLMs may vary in language modeling functions or architectures as discussed above, they can be generalized as denoising auto-encoders, which contain two procedures, ennoising and denoising. The pre-training procedure is illustrated in Figure \ref{fig:autodecoder}. 

1) Ennoising corrupts texts with arbitrary noising functions to construct training instances. The corruption scheme includes edit operations like insertion, deletion, replacement, permutation, and retrieval \citep{devlin-etal-2019-bert,lewis-etal-2020-bart,xu2021dialogue,wang2019structbert,guu2020realm}. For example, masked language modeling (MLM) \citep{devlin-etal-2019-bert} replaces some input tokens in a sentence with a special symbol. BART uses token deletion, text infilling, and sentence permutation for corruption \citep{lewis2020bart}. 

2) Denoising enables a language model to predict missing or otherwise corrupted tokens in the input sequences. Recent studies focus on designing improved language modeling functions to mitigate discrepancies between the pre-training phase and the fine-tuning phase. \citet{yang2019xlnet} reformulates MLM in XLNet by restoring the permuted tokens in factorization order, such that the input sequence is autoregressively generated after permutation. In addition, using synonyms for the masking purpose \citep{cui2020revisiting} and simple pre-training objectives based on token-level classification tasks \citep{yamaguchi2021frustratingly} have also proved effective as an MLM alternative.

Most of the existing studies of PrLMs fall into the scope of either investigating better ennoising operations or more effective denoising strategies. They treat training instances equally throughout the training process. Little attention is paid to the individual contribution of those instances. In standard MLM ennoising, randomly masking different tokens would lead to different degrees of corruption that may, therefore, cause different levels of difficulty in sentence restoration in denoising (as shown in Figure \ref{fig:autodecoder}) and thus increase the uncertainty in restoring the original sentence structure during the denoising process. For example, if ``not'' is masked, the corrupted sentence tends to have a contrary meaning.  

In this work, we are motivated to estimate the complexity of restoring the original sentences from corrupted ones in language model pre-training, to provide explicit regularization signals to encourage more effective and robust pre-training. Our approach includes two sides of penalty: 1) ennoising corruption penalty that measures the distribution disparity between the corrupted sentence and the original sentence, to measure the corruption degree in the ennoising process; 2) denoising prediction penalty that measures the distribution difference between the restored sequence and the original sentence to measure the sentence-level prediction confidence in the denoising counterpart. Experiments show that language models trained with our regularization terms can yield better performance and become more robust against adversarial attacks.

\section{Related Work}
Training powerful large-scale language models on a large unlabeled corpus with self-supervised objectives has attracted lots of attention, which commonly work in two procedures of ennoising and denoising. The most representative task for pre-training is MLM, which is introduced in \citet{devlin-etal-2019-bert} to pre-train a bidirectional BERT. A spectrum of ennoising extensions has been proposed to enhance MLM further and alleviate the potential drawbacks, which fall into two categories: 1) mask units and 2) noising scheme. Mask units correspond to the language modeling units that serve as knowledge carriers in different granularity. The variants focusing on mask units include the standard subword masking \citep{devlin-etal-2019-bert}, span masking \citep{joshi-etal-2020-spanbert}, and $n$-gram masking \citep{levine2021pmimasking,li-zhao-2021-pre}. For noising scheme, BART \citep{lewis2020bart} corrupts text with
arbitrary noising functions, including token deletion, text infilling, sentence permutation, in conjunction with MLM. UniLM~\cite{dong2019unified} extends the mask prediction to generation tasks by adding the auto-regressive objectives.
XLNet~\cite{yang2019xlnet} proposes the permuted language modeling to learn the dependencies among the masked tokens. MacBERT \citep{cui2020revisiting} suggests using similar words for the masking purpose. \citet{yamaguchi2021frustratingly} also investigates simple pre-training objectives based on token-level classification tasks as replacements of MLM, which are often computationally cheaper and result in comparable performance to MLM. In addition, ELECTRA ~\cite{clark2019electra} proposes a novel training objective called replaced token detection, which is defined over all input tokens.  

Although the above studies have an adequate investigation to reduce the mismatch between pre-training and fine-tuning tasks, an essential problem of the common denoising mechanism lacks attention. The construction of training examples based on ennoising operations would cause the break of sentence structure, either for replacement, addition, or deletion-based noising functions. In extreme cases, the destruction would lead to completely different sentences, making it difficult for the model to predict the corrupted tokens. Therefore, in this work, we propose to enhance the pre-training quality by using instance regularization (IR) terms to estimate the restoration complexity from both sides of ennoising and denoising aspects.

The proposed approach is partially related to some prior studies of hardness measurement in training deep learning models \citep{lin2017focal,kalantidis2020hard,hao2021learning}, whose focus is to guide the model to pay special attention to hard examples and prevent the vast number of easy negatives from overwhelming the training process. In contrast to optimizing the training process by heuristically finding the hard negatives, this work does not need to distinguish hard examples from ordinary ones, but measures the corruption degree between the masked sentence and the original sentence instead, and uses the degree as the explicit training signals.

\begin{figure*}
	\centering
	\includegraphics[width=0.98\textwidth]{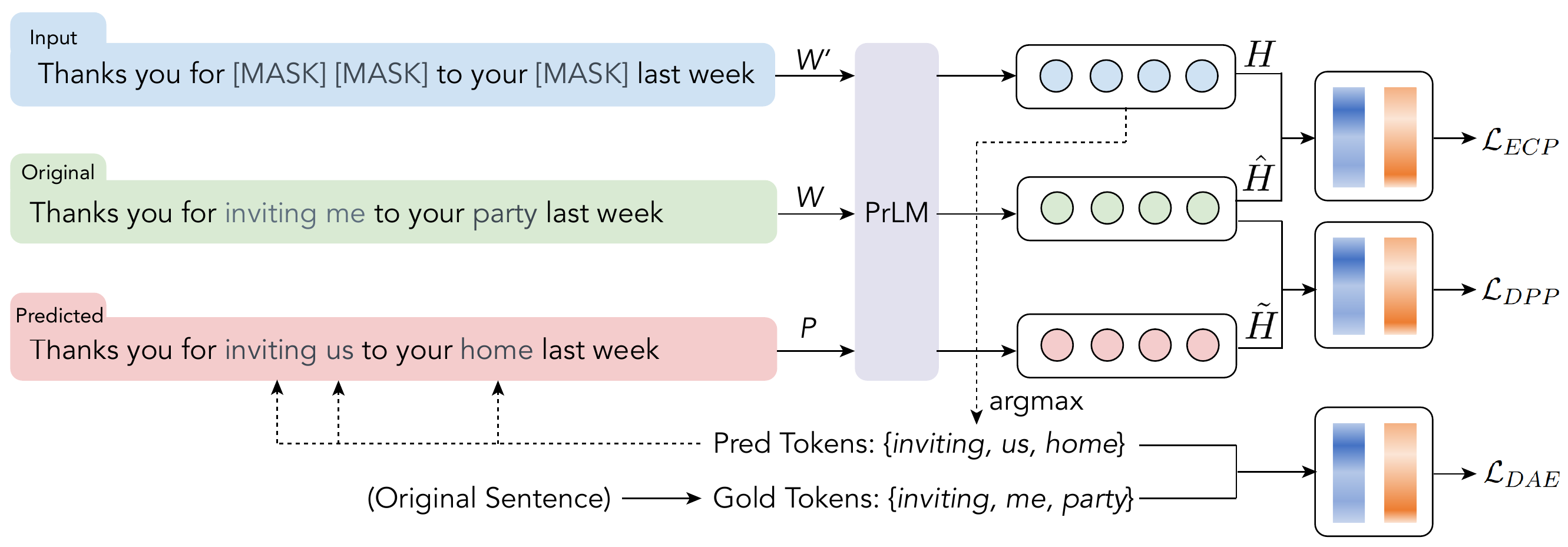}
	\caption{Overview of the procedure for the instance regularization approach, which estimates the corruption degree in the ennoising data construction process and the  prediction confidence in the denoising counterpart.}
	\label{fig:overview}
\end{figure*}

\section{Methodology}

This section will start by formulating the ennoising and denoising processes for building PrLMs and then introduce our instance regularization approach to estimate the restoration complexity in both ennoising and denoising views.

\subsection{Preliminary: Denoising Auto-Encoders}
The training procedure for discriminative language models includes ennoising and denoising processes, as described below. For the sake of simplicity, we take the widely-used MLM as a typical example to describe the ennoising process. 

\paragraph{Ennoising} Given a sentence $W = \{w_1, w_2, \dots, w_n\}$ with $n$ tokens,\footnote{We assume that $W$ has already been tokenized into a sequence of subwords.} we randomly mask some percentage of the input tokens with a special mask symbol $\texttt{[MASK]}$ and then predict those masked
tokens. Suppose that there are $m$ tokens replaced by the mask symbol. Let $\mathcal{D} = \{k_1, k_2, \dots, k_m\}$ denote set of masked positions, we have $W'$ as the masked sentence and $M = \{w_{k_1}, w_{k_2}, \dots, w_{k_m}\}$ are the masked tokens. In the following part, we use $w_{k}$ to denote each masked token for simplicity.

\paragraph{Denoising} In the denoising process, a language model is trained to predict the masked token based on the context. $W'$ is fed into the PrLM to obtain the contextual representations from the last Transformer layer, which is denoted as $H$.

\paragraph{Training} The training objective is to maximize the following objective:
\begin{equation}
    \mathcal{L}_{DAE} = - \frac{1}{m} \sum^{|D|}\limits_{k \in \mathcal{D}} \log p(w_k\mid W').
\end{equation}

\subsection{Instance Regularization}
In this part, we will introduce our instance regularization approach, which
involves two sides: corruption degree in the ennoising data construction process and the sequence-level prediction confidence in the denoising counterpart. 

During denoising, the PrLM trained by MLM is required to predict the original masked tokens $w_k$ given the hidden states $H$ of the corrupted input $W'$. Let $w'_k$ denote the predicted tokens, we replace the mask symbols by filling $w'_k$ back to $W'$. As a result, we have the predicted sequence, denoted as $P = \{p_1, p_2, \dots, p_n\}$, where the tokens in positions of $\mathcal{D}$ are predicted ones; otherwise, they are the same as the originals ones in $W$. 

Obviously, the corruption would break the sentence structure and easily cause the semantic deviation of sentence representations. According to our observation, the hidden states would vary dramatically before and after the token corruption -- similar findings were also observed in \citet{wang2021textflint} that small disturbance can inveigle PrLMs into making false predictions. In a more general perspective, replacing a modest percentage of tokens may result in a totally different sentence, let alone imperceptible disturbance as used for textual attacks. 

Therefore, we propose two approaches called ennoising corruption penalty (ECP) and denoising prediction penalty (DPP) as the regularization terms in the training process to alleviate the issue. Figure \ref{fig:overview} overviews the overall procedure. ECP measures the semantic shift from the original sentence to the corrupted one as an explicit signal to help the model distinguish easy and hard examples and learn with different weights, which can be seen as instance weighting compared with MLM. As the complement, DPP measures the sequence-level semantic distance between the predicted and original sentence to supplement the rough token-level matching of MLM, thus transforming the token prediction task to sequence matching to pay more attention to sentence-level semantics.

Both methods are used for estimating the difficulty of restoring the whole sequence from the corrupted ones, either in the role of the front-end ennoising or back-end denoising. Larger values of the estimation indicate larger semantic shifts. 

Here we go back to the formulation in MLM. As shown in Figure \ref{fig:overview}, given the original sequence $W$, the masked sequence $W'$, and the predicted sequence $P$, we obtain the contextualized representation from PrLM. Note that we already have the contextualized representation $H$ for the input sequence $W'$ in the vanilla MLM training. Similarly, we feed $W$ and $P$ to the PrLM, and the corresponding hidden states are written as $\hat{H}$ and $\tilde{H}$, respectively. Then, $H$, $\hat{H}$ and $\tilde{H}$ are leveraged as the elements for the corruption agreement and semantic agreement.

\paragraph{Ennoising Corruption Penalty} After we get the distributions $H$ and $\hat{H}$, we measure the extent of the corruption degree after ennoising by calculating the distribution difference between the masked and the original representations after  normalization:

\begin{equation}
    \centering
    \begin{aligned}
    \mathcal{L}_{ECP} = D_{KL}(H, \hat{H}),
    \end{aligned}
\end{equation}
where $KL$ refers to Kullback–Leibler (KL) divergence. Concretely, we apply softmax on the two matrices along the hidden dimensions to have two distributions. Then, we calculate KL divergence between the two distributions for each position in each sentence. Intuitively, higher $\mathcal{L}_{ECP}$ means the corruption is more severe, so is the gap between 
the ennoised instance and denoised prediction. Therefore, the model is supposed to update the gradient more significantly for those ``harder'' training instances. 

\paragraph{Denoising Prediction Penalty} In the denoising language modeling, the model would yield reasonable predictions but be discriminated as wrong predictions because such predictions do not match the single gold token for each training case using token-level cross-entropy. Therefore, we estimate the semantic agreement between the predicted sequence and the original gold sequence, by guiding the probability distribution of model predictions $\tilde{H}$ to match the expected probability distribution $\hat{H}$, we have:

\begin{equation}
    \mathcal{L}_{DPP} = D_{KL}(\tilde{H}, \hat{H}),
\end{equation}
where $\mathcal{L}_{DPP}$ is applied as the degree to reflect the sentence level semantic mismatch. 

The semantic agreement method works as a means of soft regularization to capture the sequence-level similarity as a supplement to the standard hard token-level matching in cross-entropy.

In language model pre-training, we minimize $\mathcal{L}_{H}$ and $\mathcal{L}_{S}$. Thus, the loss function is written as 
\begin{equation}
    \mathcal{L} = \mathcal{L}_{DAE} + \mathcal{L}_{ECP} + \mathcal{L}_{DPP}.\label{eq:loss}
\end{equation}

\section{Experiments}
\subsection{Setup}
To verify the effectiveness of the proposed methods, we conduct pre-training experiments and fine-tune the pre-trained models on downstream tasks. All codes are implemented using PyTorch \citep{paszke2017automatic}.\footnote{Our codes and models will be publicly available.} The experiments are run on 8 NVIDIA GeForce RTX 3090 GPUs.
\paragraph{Pre-training} We employ BERT and ELECTRA as the backbone PrLMs and implement our methods during the pre-training. For pre-training corpus, we use English Wikipedia corpus and BookCorpus \citep{zhu2015aligning} following BERT \citep{devlin-etal-2019-bert}. As suggested in \citet{liu2019roberta}, we do not use the next sentence prediction (NSP) objective as used in \citet{devlin-etal-2019-bert}, but only use MLM as the baseline language modeling objective, with a masked ratio of 15\%. After masking, 80\% of the masked positions are replaced with $\texttt{[MASK]}$, 10\% are replaced by randomly sampled words, and the remaining 10\% are kept unchanged. We set the maximum length of the input sequence to 512, and the learning rates are 3e-5. We pre-train the base and large models for 100$k$ steps using the pre-trained weights of the public BERT and ELECTRA models as initialization. The baselines are trained to the same steps for a fair comparison. To keep the simplicity like BERT training, following \citet{li2020task}, we discard the generator in ELECTRA models and use the discriminator in the same way as BERT, with a classification layer to predict the corrupted tokens. 

\paragraph{Fine-tuning} We use an initial learning rate in \{8e-6, 1e-5, 2e-5, 3e-5\} with warm-up rate of 0.1 and L2 weight decay of 0.01. The batch size is selected in \{16, 24, 32\}. The maximum number of epochs is set in [2, 5] depending on tasks. Texts are tokenized with a maximum length of 384 for SQuAD and 512 for other tasks. Hyper-parameters were selected using the development set.

\subsection{Tasks and Datasets}
For evaluation, we fine-tune the pre-trained models on GLUE (General Language Understanding Evaluation) \citep{wang-etal-2018-glue} and the popular Stanford Question Answering Dataset (SQuAD) \citep{Rajpurkar2016SQuAD} to evaluate the performance of the pre-trained models. The concerned tasks involve natural language inference, semantic similarity, text classification, and machine reading comprehension (MRC).

\paragraph{Natural Language Inference}
Natural Language Inference involves reading a pair of sentences and judging the relationship between their meanings, such as entailment, neutral and contradiction. We evaluate on three diverse datasets, including Multi-Genre Natural Language Inference (MNLI) \cite{nangia2017repeval}, Question Natural Language Inference (QNLI) \cite{Rajpurkar2016SQuAD} and Recognizing Textual Entailment (RTE) \cite{bentivogli2009fifth}.

\paragraph{Semantic Similarity}
Semantic similarity tasks aim to predict whether two sentences are semantically equivalent or not. The challenge lies in recognizing rephrasing of
concepts, understanding negation, and handling syntactic ambiguity. Three datasets are used, including Microsoft Paraphrase corpus (MRPC) \cite{dolan2005automatically}, Quora Question Pairs (QQP) dataset \cite{chen2018quora} and Semantic Textual Similarity benchmark (STS-B) \cite{cer2017semeval}.

\paragraph{Classification}
The Corpus of Linguistic Acceptability (CoLA) \cite{warstadt2018neural} is used to predict whether an English sentence is linguistically acceptable or not. The Stanford Sentiment Treebank (SST-2) \cite{socher2013recursive} provides a dataset for sentiment classification that needs to determine whether the sentiment of a sentence extracted from movie reviews is positive or negative.

\paragraph{Reading Comprehension}
As a widely used MRC benchmark dataset, SQuAD \citep{Rajpurkar2016SQuAD} is a reading
comprehension dataset that requires the machine to extract the answer span given a document along with a question. We select the v1.1 version to keep the focus on the performance of pure span extraction performance. Two official metrics are used to evaluate the model performance: Exact Match (EM) and a softer metric F1 score, which measures the average overlap between the prediction and ground truth answer at the token level. 

\begin{table*}[htb]
    \centering
    \setlength{\tabcolsep}{5pt}
    \begin{tabular}{l c c c c c c c c c}
    \toprule
    \multirow{2}{*}{\textbf{Model}} & \textbf{CoLA} & \textbf{SST-2} & \textbf{MRPC} & \textbf{STS-B} & \textbf{QQP} & \textbf{MNLI} & \textbf{QNLI} & \textbf{RTE} & \textbf{Average} \\
    & \textit{Mcc}   & \textit{Acc}   & \textit{Acc}   & \textit{Spear}   & \textit{Acc}   & \textit{Acc}   & \textit{Acc}   & \textit{Acc} & - \\
    \midrule    
    \multicolumn{10}{l}{\textit{{Results on the development  sets}}} \\
    \midrule
    BERT$_{\rm base}$ & 59.32&	92.32 &	87.25 &	87.36&	90.78&	84.75&	91.42&	65.34&	82.32 \\
    BERT-IR$_{\rm base}$ & 61.39&	93.46 &87.50&	89.05 &	90.90 &	85.28 &	91.84 	&68.95&	83.52 \\
    BERT$_{\rm large}$&  62.45 &	93.58  &	88.24 &	90.48 &	91.45 
 &	87.20 &	92.37 &	74.01  & 84.97 
 \\
    BERT-IR$_{\rm large}$ & 64.07 & 	94.27 & 	88.73 & 	90.57 & 	91.55  & 	87.35 & 	92.71 & 	75.09 & 	85.54  \\
    \hdashline
    ELECTRA$_{\rm base}$ & 65.53 &	94.95 &	88.97 &	89.96 &	91.24 &	88.45 &	92.53 &	77.62 &	86.16 
 \\
    ELECTRA-IR$_{\rm base}$ & 68.95 & 	95.30 & 	90.44 & 	90.52 & 	91.40 & 	88.66 & 	93.04 & 	79.06 & 	87.17 
 \\
    ELECTRA$_{\rm large}$ & 70.41 & 	96.79 & 	89.22 & 	91.92 & 	92.07 & 	90.26 & 	94.40 & 	85.92 & 	88.87
\\
    ELECTRA-IR$_{\rm large}$ &72.09 &	97.48 &	91.18 &	92.03 &	92.27 &	90.55&	94.64 &	87.36 	&89.70 
\\
    \midrule    
    \multicolumn{10}{l}{\textit{{Results on the test sets}}} \\
    \midrule
    BERT$_{\rm{base}}$ &  52.1 & 93.5 & 84.8 & 85.8 & 89.2 & 84.6 & 90.5 & 66.4 & 80.9   \\
    BERT-IR$_{\rm base}$ & 54.1 & 	93.9 & 	84.9 & 	86.6 & 	89.1 & 	85.3 & 	91.1 & 	71.2 & 	82.0\\
    BERT$_{\rm{large}}$ &  60.5 & 94.9 & 85.4 & 86.5 & 89.3 & 86.7 & 92.7 & 70.1 & 83.3 \\
    BERT-IR$_{\rm large}$ & 61.7 & 	94.2 & 	85.7	 & 87.1	 & 89.4 & 	86.5 & 	92.9 & 	72.1 & 83.7  \\
    \hdashline
    ELECTRA$_{\rm{base}}$ &  59.7 & 93.4 & 86.7 & 87.7 & 89.1 & 85.8 & 92.7 & 73.1 & 83.5 \\
    ELECTRA-IR$_{\rm base}$ & 63.2	& 95.4	& 86.5& 	89.0	& 89.2	& 88.4	& 92.9	& 70.7	& 84.4 \\
    ELECTRA$_{\rm{large}}$ & 68.1  & 96.7  & 89.2  & 91.7  & 90.4  & 90.7  & 95.5  & 86.1  & 88.6 \\
    ELECTRA-IR$_{\rm large}$ & 70.1&	97.0	&89.8&	91.6&	90.2	&90.9&	95.8&	86.8&	89.0 \\
    \bottomrule
    \end{tabular}
    \caption{Comparisons between our proposed methods and the baseline models under on the GLUE development sets. STS-B is reported by Spearman correlation, CoLA is reported by Matthew's correlation, and the other tasks are reported by accuracy. Only one decimal place is reserved for the test results which are from the online GLUE server.}
    \label{result:GLUE}
\end{table*}

\begin{table}[htb]
    \centering
\setlength{\tabcolsep}{5.0pt}
\begin{tabular}{lcc}
    \toprule
    \textbf{Model}    & \textbf{EM Score} & \textbf{F1 Score}\\
    \midrule
    BERT$_{\rm base}$ & 80.48 &	87.77  \\
    BERT-IR$_{\rm base}$  & 81.28 &	88.38 \\
    BERT$_{\rm large}$  &83.54 &	90.26\\
    BERT-IR$_{\rm large}$  & 84.26	 & 90.92\\
    \midrule
    ELECTRA$_{\rm base}$ & 83.82 & 90.59\\
    ELECTRA-IR$_{\rm base}$ & 84.49&	91.18\\
    ELECTRA$_{\rm large}$ & 87.59 &	93.78\\
    ELECTRA-IR$_{\rm large}$ & 88.34 &	94.09\\
    \bottomrule
  \end{tabular}
  \caption{Results on the SQuAD development set. The evaluation metrics are Exact-Match (EM) and F1 scores.}\label{exp-squad}
\end{table}

\begin{table*}[htb]
    \centering
    \setlength{\tabcolsep}{5pt}
    \begin{tabular}{l c c c c c c c c c }
    \toprule
    \multirow{2}{*}{\textbf{Model}} & \textbf{CoLA} & \textbf{SST-2} & \textbf{MRPC} & \textbf{STS-B} & \textbf{QQP} & \textbf{MNLI} & \textbf{QNLI} & \textbf{RTE} & \textbf{Average} \\
    & \textit{Mcc}   & \textit{Acc}   & \textit{Acc}   & \textit{Spear}   & \textit{Acc}   & \textit{Acc}   & \textit{Acc}   & \textit{Acc} & - \\
    \midrule    
    BERT-IR$_{\rm{base}}$  & 61.39&	93.46 &87.50&	89.05 &	90.90 &	85.28 &	91.84 	&68.95&	83.52 \\
    \quad - ECP & 60.84 & 93.11 & 88.48 & 87.13 & 90.83 & 84.94 & 91.54 & 66.78 & 82.96 \\
    \quad - DPP & 59.90  & 93.23 & 87.01 & 87.43 & 90.89 & 84.70 & 91.43 & 67.87 & 82.81 \\
    \midrule
    ELECTRA-IR$_{\rm{base}}$ & 68.95 & 	95.30 & 	90.44 & 	90.52 & 	91.40 & 	88.66 & 	93.04 & 	79.06 & 	87.17 
   \\
    \quad - ECP& 67.08&	95.21 &	89.71 &	90.26 &	91.17 &	88.61 &	92.87 &	77.26 &	86.52 
 \\
    \quad - DPP & 67.75 &	95.18 &	89.21 &	90.35 &	91.28 &	88.50 &	92.75 &	76.89 &	86.49 
 \\
    \bottomrule
    \end{tabular}
    \caption{Ablation study of the proposed methods under BERT-base and ELECTRA-base on the GLUE development set. STS-B is reported by Spearman correlation, CoLA is reported by Matthew's correlation, and other tasks are reported by accuracy.}
    \label{ana:abl}
\end{table*}

\begin{table*}[htb]
    \centering
    \setlength{\tabcolsep}{6.8pt}
    \begin{tabular}{l c c c c c c c c c }
    \toprule
    \multirow{2}{*}{\textbf{Model}} & \textbf{CoLA} & \textbf{SST-2} & \textbf{MRPC} & \textbf{STS-B} & \textbf{QQP} & \textbf{MNLI} & \textbf{QNLI} & \textbf{RTE} & \textbf{Average} \\
    & \textit{Mcc}   & \textit{Acc}   & \textit{Acc}   & \textit{Spear}   & \textit{Acc}   & \textit{Acc}   & \textit{Acc}   & \textit{Acc} & - \\
    \midrule    
    BERT$_{\rm{base}}$ & 30.88 & 87.84 & 	74.75 &	76.06 &	87.77 & 75.70 &	84.24 &	56.68 &	71.74 \\
    BERT-IR$_{\rm{base}}$ & 33.06 & 89.56 &	76.47 & 76.64 & 88.08 &	76.23 & 85.03 &	59.57 &	73.08 \\
    \bottomrule
    \end{tabular}
    \caption{Results of training BERT$_{\rm{base}}$ and BERT-IR$_{\rm{base}}$ from scratch.}
    \label{ana:scr}
\end{table*}

\begin{figure*}[htb]
   \centering
   \subfigure{
   \begin{minipage}[b]{0.48\linewidth}
\setlength{\abovecaptionskip}{0pt}
\begin{center}
\pgfplotsset{height=5.5cm,width=8cm,compat=1.14,every axis/.append style={thick},every axis legend/.append style={ at={(0.95,0.95)}},every tick label/.append style={font=\small},legend columns=3 row=2} \begin{tikzpicture} \tikzset{every node}=[font=\small]

\begin{axis} [width=8.2cm,enlargelimits=0.13,legend pos=north west,legend cell align={left}, xticklabels={12.5$k$, 25.0$k$, 50$k$, 80$k$, 100$k$}, axis y line*=left, axis x line*=left, xtick={0,1,2,3,4}, x tick label style={rotate=0},
ylabel={Acc}, 
ymin=58,ymax=63,
  ylabel style={align=left},xlabel={CoLA},font=\small]

\addplot+[smooth, mark=*,mark size=1.2pt,mark options={solid,mark=*,red}, color=red] coordinates {(0, 58.8) (1, 58.6) (2, 59.0) (3, 59.1) (4, 59.3)};
\addlegendentry{\scriptsize BERT$_{\rm{base}}$}

\addplot+ [smooth, mark=star,mark size=1.2pt,mark options={mark color=brown}, color=brown] coordinates { (0, 59.3)  (1, 60.3) (2, 60.6) (3, 60.8) (4, 61.4)};
\addlegendentry{\scriptsize BERT-IR$_{\rm{base}}$}


\end{axis}
\end{tikzpicture}

\end{center}
   \end{minipage}
   }
   \subfigure{
   \begin{minipage}[b]{0.48\linewidth}
\setlength{\abovecaptionskip}{0pt}
\begin{center}

\pgfplotsset{height=5.5cm,width=8cm,compat=1.14,every axis/.append style={thick},every tick label/.append style={font=\small},every axis legend/.append style={ at={(0.95,0.95)}},legend columns=3 row=2} 

\begin{tikzpicture} \tikzset{every node}=[font=\small] 
\begin{axis} [width=8.2cm,enlargelimits=0.13,legend pos=north west,legend cell align={left}, xticklabels={12.5$k$, 25.0$k$, 50$k$, 80$k$, 100$k$}, axis y line*=left, axis x line*=left, xtick={0,1,2,3,4}, x tick label style={rotate=0},
ylabel={Acc}, 
ymin=84.5,ymax=85.8,
  ylabel style={align=left},xlabel={MNLI},font=\small]

\addplot+[smooth, mark=*,mark size=1.2pt,mark options={solid,mark=*,red}, color=red] coordinates {(0, 84.72) (1, 84.67) (2, 84.70) (3, 84.73) (4, 84.75)};
\addlegendentry{\scriptsize BERT$_{\rm{base}}$}

\addplot+ [smooth, mark=star,mark size=1.2pt,mark options={mark color=brown}, color=brown] coordinates { (0, 84.89)  (1, 85.01) (2, 85.16) (3, 85.21) (4, 85.28)};
\addlegendentry{\scriptsize BERT-IR$_{\rm{base}}$}

\end{axis}
\end{tikzpicture}
\end{center}
   \end{minipage}
   }
    \caption{The performance (accuracy) of different training steps of BERT$_{\rm{base}}$ and BERT-IR$_{\rm{base}}$ on the CoLA and MNLI development sets.}
    \label{fig:efficiency}
\end{figure*}
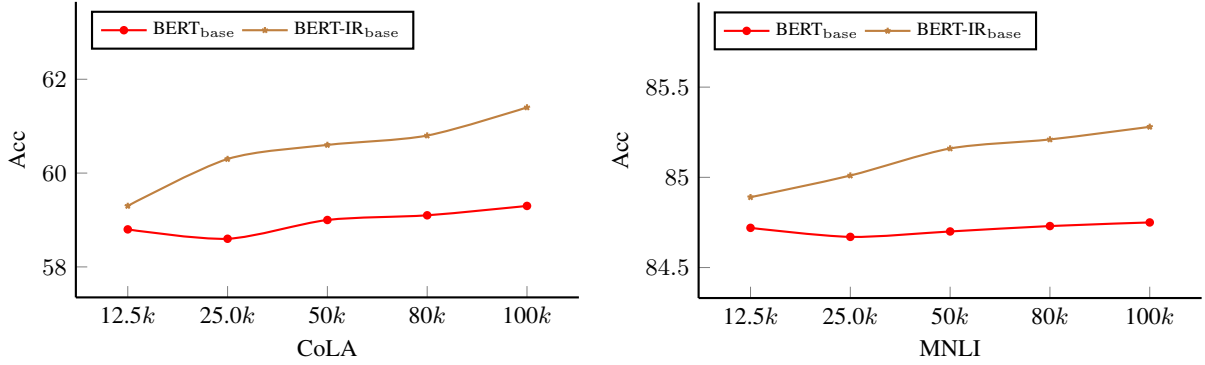

\subsection{Main Results}

Table \ref{result:GLUE} presents the results of our methods and baselines under the same pre-training settings on the GLUE development sets. We see that our method achieves consistent performance gains over both BERT and ELECTRA baselines under the base and large settings, i.e., with the increased average scores of +1.20\%/0.57\% on BERT (base/large) and +1.01\%/0.83\% on ELECTRA (base/large).\footnote{We report the results on large models just to verify the consistent advance instead of pursuing absolute scores, due to the difficulty of training larger models on a single machine with 8 NVIDIA GTX 3090 GPUs (e.g., weak convergence with small batch sizes).} The results indicate that the instance regularization approach is effective for improving the general language understanding capacity of PrLMs.  

We also show the comparisons with public methods on the GLUE test sets. For a fair comparison, we only compare with the related single models fine-tuned for a single task, without model ensembling and task-specific tricks. According to the results, we observe that our models yield consistent advances on most of the tasks compared with public BERT and ELECTRA models under both base and large sizes.
 
We further evaluate the performance of our models on the challenging SQuAD MRC task. Table \ref{exp-squad} shows the results, which indicate modest performance gains in the reading comprehension task. The results show that our method is not only effective for the sentence-level classification or regression tasks of NLU but also beneficial for passage-level reading comprehension.

\section{Analysis}
\subsection{Ablation Study}
To investigate the contribution of the internal components of the proposed IR objective, we conduct an ablation study under BERT$_{\rm base}$ and ELECTRA$_{\rm base}$ on the GLUE development set. Table \ref{ana:abl} shows the performance when removing each one of the methods. We observe that removing either ECP or DPP objective will result in performance drop generally, which verifies the effectiveness of both methods.

\begin{table}[htb]
    \centering
\setlength{\tabcolsep}{8.8pt}
\begin{tabular}{lcc}
    \toprule
    \textbf{Baseline} & \textbf{KL Divergence} & \textbf{MSE}\\
    \midrule
    82.32 &	83.52 & 83.27 \\
    \bottomrule
  \end{tabular}
  \caption{Comparison of using KL divergence and MSE to measure the distribution distances}\label{ana-distance}
\end{table}

\subsection{Comparison with other distance measures}
We apply KL divergence to measure the distance between distributions. We compare the performance for different distance measures by using mean-square error (MSE) loss. The average GLUE scores (based on BERT-base) are shown in Table \ref{ana-distance}. We see that both IR methods contribute to better performance. The results further verify the general benefits of instance regularization for pre-training no matter what the distance function is. 

\subsection{Performance in Different Training Steps}
To interpret the training effectiveness of our proposed method, we illustrate the performance of different training steps of BERT$_{\rm{base}}$ and BERT-IR$_{\rm{base}}$ on the development sets of the small-scale CoLA and the large-scale MNLI tasks, as shown in Figure \ref{fig:efficiency}. We see that the accuracy of the baselines boost slightly as the training steps increase. In contrast, our models can still yield obvious gains, which indicates our PrLM models could absorb extra beneficial signals via the newly proposed  instance regularization approach. 

\begin{table*}[htb]
\centering
\setlength{\tabcolsep}{3.2pt}
{
\begin{tabular}{lcccccc}
\toprule
\multirow{2}{*}{\textbf{Model}}
& \multicolumn{2}{c}{\textbf{\emph{Original Reference}}}
& \multicolumn{2}{c}{\textbf{\emph{SwapSynWordEmbedding}}}
& \multicolumn{2}{c}{\textbf{\emph{SwapSynWordNet}}}
\\
& EM Score & F1 Score
& EM Score & F1 Score
& EM Score & F1 Score
\\
\midrule
BERT$_{\rm base}$ & 85.33 & 88.78 & 84.67 ($\downarrow$0.67) & 87.67 ($\downarrow$1.11) & 81.67 ($\downarrow$3.67) & 85.15 ($\downarrow$3.63)  \\
BERT-IR$_{\rm base}$ & 84.33 & 87.70& 84.67 (\textbf{$\uparrow$0.33}) & 87.82 (\textbf{$\uparrow$0.12}) & 82.33 (\textbf{$\downarrow$2.00}) & 85.42 (\textbf{$\downarrow$2.28})\\
\midrule
ELECTRA$_{\rm base}$ & 89.00 & 90.91 & 86.67 ($\downarrow$2.33) & 88.89 ($\downarrow$2.02) & 87.00 ($\downarrow$2.00) & 89.39 ($\downarrow$1.53)  \\
ELECTRA-IR$_{\rm base}$ & 89.67 & 91.44 & 89.00 (\textbf{$\downarrow$0.67}) & 90.30 (\textbf{$\downarrow$1.14}) & 89.00 (\textbf{$\downarrow$0.67}) & 91.03 (\textbf{$\downarrow$0.41})  \\
\bottomrule
\end{tabular}
}
\caption{Robustness evaluation on the SQuAD dataset. \textit{Original} means the results of original dataset sampled from the SQuAD v1.1 development set by TextFlint \citep{wang2021textflint}, and \textit{Swap}. indicates the transformed one. The assessed models are the ${\rm base}$ models from Table \ref{exp-squad}. In this analysis, the lower performance drop means the better.}
\label{tab:robust}
\end{table*}

\subsection{Convergence Speed}
\begin{figure}[htb]
	\centering
	\includegraphics[width=0.46\textwidth]{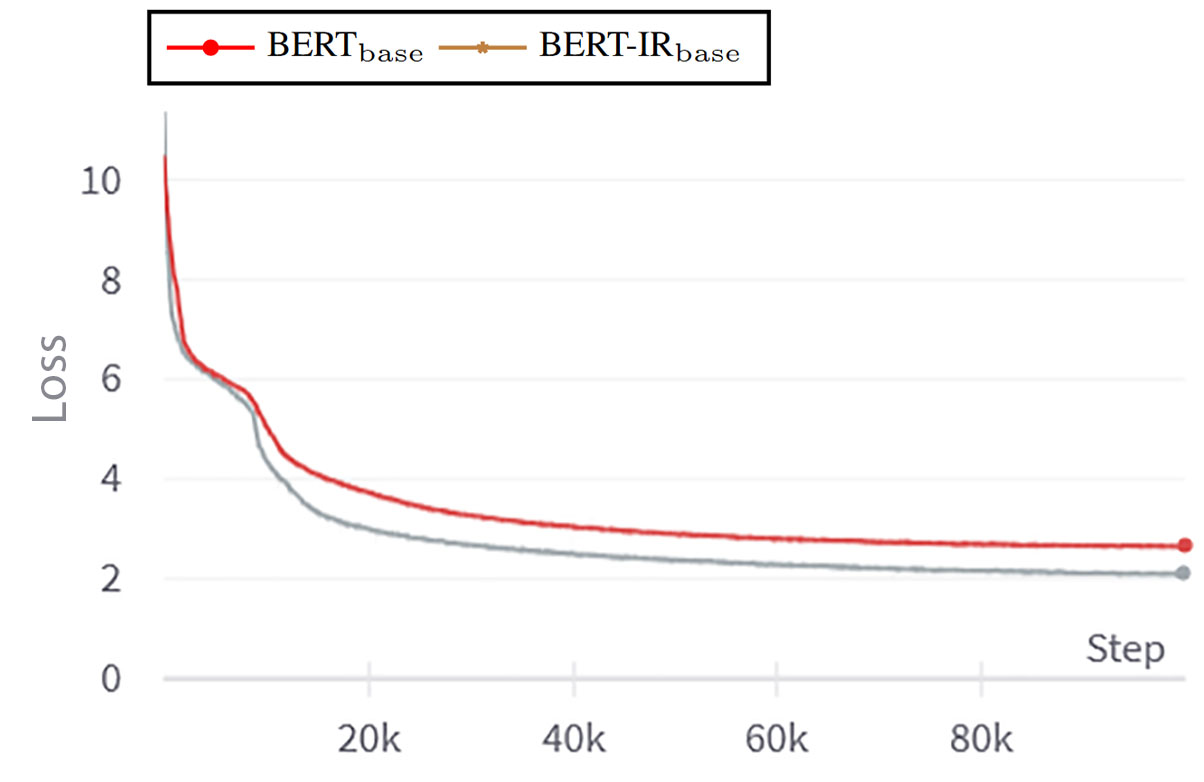}
	\caption{Training curve of the BERT-based models.}
	\label{fig:curve}
\end{figure}

Figure \ref{fig:curve} shows the training curve of the BERT$_{\rm base}$ and BERT-IR$_{\rm base}$ models when training from scratch.\footnote{For clear observation, we pre-train the baseline and our model from scratch instead of continuous training, because the checkpoints used for continuing training have already converged under a small loss, making it hard to interpret the convergence.} We observe that the absolute values of our approaches are relatively higher than the baselines at the very beginning. The reason is that our loss function is composed of three elements as formalized in Eq. \ref{eq:loss}. However, our model converges quickly. The loss of BERT-IR$_{\rm base}$ falls below the baseline when the training goes on, and the slope of our curve is obviously larger than that of the baseline. In addition, we also evaluate the baseline and our model trained from scratch (Table \ref{ana:scr}), which achieve the average accuracy of 71.74\% and 73.08\% (+1.3\%) on the GLUE datasets, respectively. The analysis above  indicates that the PrLM model trained with our approach could absorb extra knowledge via the newly proposed instance regularization approach, and it would benefit the training of the vanilla masked language modeling as well. 

\subsection{Training Cost}\label{sec:cost}
Since the calculation of the regularization terms involves two forward passes of input sequences, we further investigate the influence of the training cost. Analysis shows that our model is efficient in training speed and parameter size. Taking the BERT-based model for example, the training time of BERT-IR$_{\rm base}$ and BERT$_{\rm base}$ baseline for 200K steps is 67h/74h (only 10\% increase) with the same hyper-parameters on base models. The memory cost also keeps basically the same scale as the baseline since the regularization does not necessarily require extra gradient backpropagation. 

\subsection{Robustness Against Synonym-Based Adversarial Attacks}
The semantic agreement in IR measures the consistency between similar sentences, which may improve our model's robustness. To verify the hypothesis, we evaluate our models in Table \ref{exp-squad} with synonym-based adversarial examples derived from the SQuAD v1.1 development set. The examples are generated by a robustness evaluation platform TextFlint \citep{wang2021textflint}, using \textit{SwapSynEmbedding} and \textit{SwapSynWordNet}, which transform an input by replacing its words with synonyms provided by GloVe embeddings \citep{pennington2014glove} or WordNet \citep{miller1998wordnet}, respectively.

The results are shown in Table \ref{tab:robust}, from which we observe that the adversarial attacks can lead to an obvious performance drop of the baseline models, i.e., 3.67(EM)\%/3.63(F1)\% of BERT$_{\rm base}$ on \textit{SwapSynWordNet}. In contrast, our models perform less sensitively against the adversarial examples, and our BERT-IR$_{\rm base}$ even yields an increase of scores in \textit{SwapSynEmbedding} attack. The results indicate that the regularization helps the model to resist synonym-based adversarial attacks with less performance degradation.


\section{Conclusion}
In this paper, we study the instance-aware contribution estimation from the ennoising and denoising processes in discriminative language model pre-training, motivated by the observation that the quality of denoising has to be subject to the complexity of the constructed training data from ennoising. The estimation is decomposed into ennoising corruption penalty and denoising prediction penalty, which are used as regularization terms for language model pre-training. Experiments show that language models trained with our regularization terms can yield improved performance on downstream tasks, with better robustness against adversarial attacks. In addition, the training efficiency can be improved as well, without severe costs of computation resources and training speed. We hope our work could facilitate related studies to improve training quality while keeping a lightweight model size.

\section{Limitations}
We acknowledge that the major limitation of the proposed method is additional computation compared with the vanilla language models because the calculation of the regularization terms involves two forward passes of input sequences. As discussed in Section \ref{sec:cost}, the training time of BERT-IR$_{\rm base}$ and BERT$_{\rm base}$ baseline for 200K steps is 67h/74h (about 10\% increase) with the same hyper-parameters on base models. A more efficient instance regularization method without additional training passes could be future work.

\bibliography{custom}
\bibliographystyle{acl_natbib}

\end{document}